\documentclass[twoside,11pt]{article}

\usepackage{jmlr2e}

\usepackage{graphicx}
\usepackage{caption}
\usepackage{subcaption}
\usepackage[algoruled,vlined,nofillcomment]{algorithm2e}
\usepackage{amsmath}

\usepackage{enumitem}
\usepackage{amsfonts}

\jmlrheading{1}{}{}{}{}{}

\firstpageno{1}

\begin{document}

\title{A penalisation method for batch multi-objective Bayesian optimisation with application in heat exchanger design}

\author{\name Andrei Paleyes \email ap2169@cam.ac.uk \\
       \addr Department of Computer Science and Technology, University of Cambridge
       \AND
       \name Henry B. Moss \& Victor Picheny \\
       \addr Secondmind.ai
       \AND
        \name Piotr Zulawski \& Felix Newman  \\
       \addr Reaction Engines}

\editor{}

\maketitle
\vspace{-30pt}
\begin{abstract}
We present HIghly Parallelisable Pareto Optimisation (HIPPO) --- a batch acquisition function that enables multi-objective Bayesian optimisation methods to efficiently exploit parallel processing resources. Multi-Objective Bayesian Optimisation (MOBO) is a very efficient tool for tackling expensive black-box problems.
However, most MOBO algorithms are designed as purely sequential strategies, and existing batch approaches are prohibitively expensive for all but the smallest of batch sizes. We show that by encouraging batch diversity through penalising evaluations with similar predicted objective values, HIPPO is able to cheaply build large batches of informative points. Our extensive experimental validation demonstrates that HIPPO is at least as efficient as existing alternatives whilst incurring an order of magnitude lower computational overhead and scaling easily to batch sizes considerably higher than currently supported in the literature. Additionally, we demonstrate the application of HIPPO to a challenging heat exchanger design problem, stressing the real-world utility of our highly parallelisable approach to MOBO.
\end{abstract}

\begin{keywords}
  Multi-objective optimisation, Bayesian optimisation, Batch optimisation
\end{keywords}

\vspace{-10pt}
\section{Introduction}
\vspace{-5pt}
Bayesian Optimisation (BO, \citealp{shahriari2015taking}) is a well-established strategy to tackle expensive-to-evaluate black-box optimisation problems,
and in particular the subclass of problems for which one seeks a trade-off between several equally important but competing objectives.
In this so-called \textit{Multi-Objective Optimisation} (MOO) context an improvement in one metric may harm another, and the objective is to find a set of points corresponding to optimal trade-offs, called the \textit{Pareto set}\footnote{Its image in the objective space being referred to as the \textit{Pareto front}.}. Numerous problems in science and industry are framed in that way \citep{park2018multi, Nahvi2019MultiobjectiveBO, avent2020automatic}.


An increasingly popular approach for speeding up MOO is to exploit parallel processing resources, i.e. allowing multiple evaluations to be queried in batches. Access to such parallel resources is common in life science and industry applications, where large batches of experiments can be ran concurrently, or in machine learning, where cloud computing resources can be used to parallelise computation. Although the classical formulation of BO is primarily sequential, extending BO ideas to this batch settings has been an active area of research and there exists many efficient methods for single-objective BO that allow arbitrary degrees of parallelisation. However, for multi-objective BO, most currently available approaches for parallel optimisation allow only modest batch sizes, and their performance drops significantly as the optimisation loop progresses or as the batch size increases. 

In this work, we introduce HIPPO (HIghly Parallelisable Pareto Optimisation), a computationally efficient method for collecting large batches of points within multi-objective BO. Inspired by the penalisation method of \cite{Gonzlez2016BatchBO}, we propose an iterative batch acquisition strategy with multiplicative distance-based penalisation. Compared to existing alternatives, HIPPO incurs a significantly smaller computational overhead when collecting each new batch point and is agnostic to the surrogate model of the objectives.

\vspace{-10pt}
\section{Related Work}\label{sec:related-work}
\vspace{-5pt}
For non-batch BO, Expected Hypervolume Improvement (EHVI, \citealp{emmerich2005single}) is a popular acquisition function for MOO. It aims to find the point which induces a Pareto front with maximimal expected improvement in hypervolume. Under the efficient implementation suggested by \cite{yang2019efficient}, EHVI is currently considered a gold standard, and forms the basis of our HIPPO acquisition function. 

There are three primary ways to achieve batch BO. Firstly, acquisition functions can be defined over batches of points, for instance the (single objective) multi-point expected improvement \citep{chevalier2013fast} with recent extensions to multi-objective optimisation \citep{Daulton2020DifferentiableEH, daulton2021parallel}. These approaches measure a Monte-Carlo approximation of  the likely improvement in the Pareto front provided by candidate batches and is henceforth referred to as Batch MC. However, such approaches scale very poorly, as they usually incur a high computational cost and  acquisition optimisation over entire batches is  challenging.

An alternative approach for batch BO is to define a set of diverse acquisition functions and optimise each one independently, with Thompson sampling being the canonical example. Here, each acquisition function is a sample from the GP model \citep{vakili2021scalable}. Similar methods can also be applied to multi-objective BO, for example Diversity-Guided Efficient Multi-objective Optimisation (DGEMO, \citealp{KonakovicLukovic2020DiversityGuidedMB}) which divides the currently discovered front into sectors and  independently picks  points from each sector, or random scalarisation-based Thompson sampling \citep{paria2020flexible}. While these approaches have the advantage of trivial parallelisation, diversity of batches is only controlled indirectly, i.e. there is no guarantee that batches are diverse.

The third approach for batch BO, often referred to as ``greedy'', constructs a batch by sequentially adding the maximiser of an acquisition function to the batch, then updating the acquisition function based on the location of the newly added point. 
As examples for single objectives: Kriging Believer (KB, \citealp{Ginsbourger2010KrigingIW}) conditions the GP on fake observations at the points already added to the batch (effectively modifying the posterior distribution and hence also the acquisition function); local penalisation \citep[LP,][]{Gonzlez2016BatchBO,Alvi2019AsynchronousBB} applies multiplicative penalisation terms based on assumed Lipschitz continuity of the objective function; and GIBBON \citep{Moss2021GIBBONGI} sequentially accumulates points expected to provide large reductions in our current uncertainty of the function's maximum value. Among greedy approaches, KB can be used in the multi-objective setting, but only with GP regression models and for very small batch sizes. Recent work has also applied entropy-based method to multi-objective optimization, however, these do not scale to provide batches containing more that a handful of points and so are not considered in this paper \citep{qing2022text}. Despite convincing performance in the single objective case, no penalisation methods have been proposed yet for MOO: this is the purpose of HIPPO.

\vspace{-10pt}
\section{HIPPO: Penalisation for Multi-objective Optimisation}
\vspace{-5pt}
We now present the primary contribution of this work ---  a penalisation method for effective batch multi-objective optimisation of a black-box multi-output function. Mathematically, we wish to discover the Pareto front between the $k$ outputs of a function

\begin{equation}
f(x) = (f_1(x), f_2(x), ..., f_k(x)), x \in \mathcal{X} \subset \mathcal{R}^n, f_i(x) \in \mathcal{R},
\end{equation}
where $n$ is the dimensionality of the input space, and $k$ --- of the output space.

We built HIPPO based on the following principles, which are also shared with the single-objective LP approaches: 1) simplicity to ensure scalability with batch size, 2) model-agnosticity, 3) acquisition function agnosticity.
Existing LP approaches directly encourage batch diversity through a repulsion mechanism in the input space, where the strength of this repulsion reduces in the vicinity of the current optimum (points in the batch can be close to each other \textit{only} if we believe they are close to optimal).
However, such approaches are not applicable to multi-objective problems where the goal is to discover a set rather than a single point. Ensuring diversity in the input space may not achieve diversity along the Pareto front, as distant parts of the input domain might correspond to the same area of the Pareto front. 



Instead, we propose a penalty based on distances in the objective space, i.e. a candidate point will have its acquisition function penalised if its predictive objectives are close to the predicted objectives of a point already added to the batch. This may ensure the generation of batches of points that have high diversity along the Pareto front.
Specifically, we consider a warped Mahalanobis distance defined between the posterior distributions of our surrogate models at the candidates points. Note that if we follow a common assumption of independence between Gaussian process models of each objective \citep{Svenson2016MultiobjectiveOO}, the Mahalanobis distance $d:\mathcal{X}\times\mathcal{X} \rightarrow \mathcal{R} $ becomes:
\begin{equation}
	d(x, y) = \sqrt{\sum_{i=1}^{n} \frac{(\mu(x[i]) - \mu(y[i])^2}{(\sigma(y[i]))^2}},
\end{equation}
 where $\mu(x)\in\mathcal{R}^k$ and $\sigma(x)^2\in\mathcal{R}$ are the mean and variance of our surrogate models at location $x$, respectively. Intuitively, this distance will induce a penalty that increases if both predictive means are similar; and this penalty will increase further if the prediction variance is small, i.e. if the relative position of the points is known with certainty.

Following the construction of \cite{Gonzlez2016BatchBO} (where each new point in the batch incurs a multiplicative penalty term to the acquisition), we propose the HIPPO acquisition function for the $(t+1)$st element of the batch as
\begin{equation}
	\alpha_{t+1}(x) = \alpha(x) \times \prod_{j=1}^{t} \phi(x, x_{j}^*) = \alpha(x) \times \prod_{j=1}^{t} w(d(x, x_j)),
\end{equation}
where $1 \leq t \leq b-1$ is the number of points in the batch collected so far, $b$ is the total batch size, $\phi(.,.)$ is the penalisation function, $w: \mathcal{R} -> [0, 1)$ is a smooth warping function and $\alpha:\mathcal{X}\rightarrow\mathcal{R}$ is the well-known (but non-batch) EHVI acquisition function.

\begin{figure*}[t]
	\newcommand{\scale}{0.32}
	\begin{subfigure}[b]{\scale\textwidth}
	    \centering
		\includegraphics[width=\textwidth]{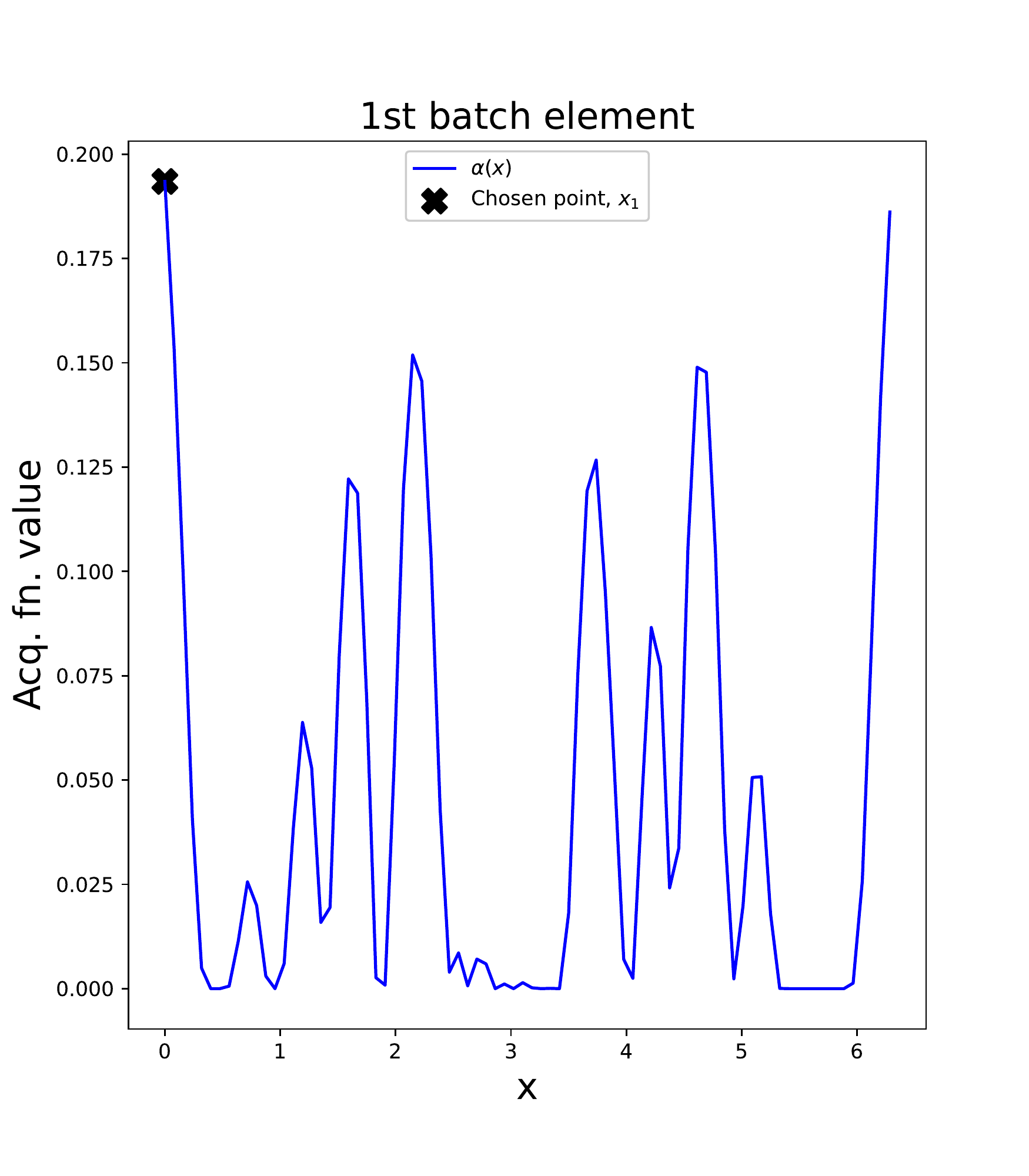}
	\end{subfigure}
	\begin{subfigure}[b]{\scale\textwidth}
		\centering
		\includegraphics[width=\textwidth]{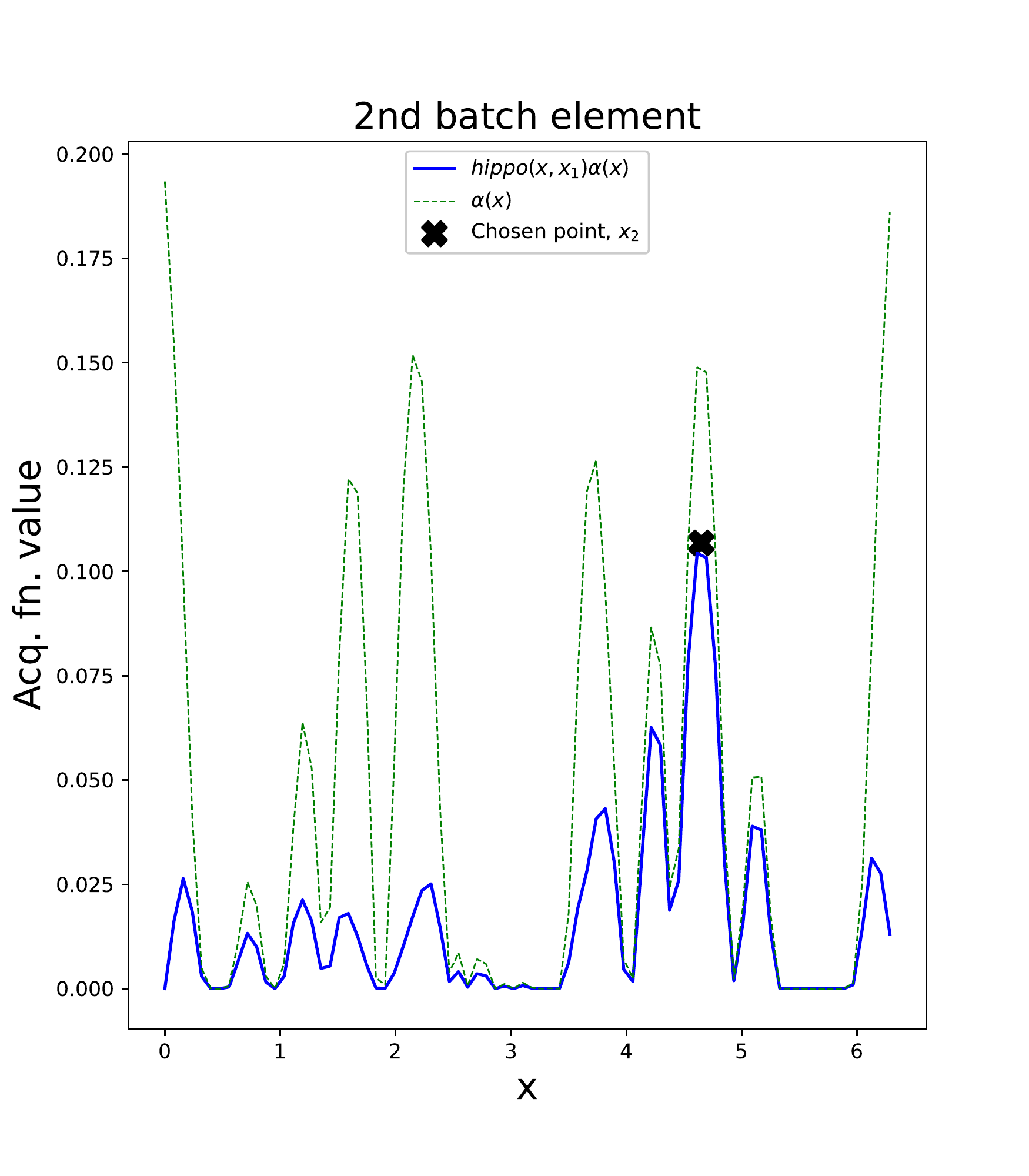}
	\end{subfigure}
	\begin{subfigure}[b]{\scale\textwidth}
	    \centering
		\includegraphics[width=\textwidth]{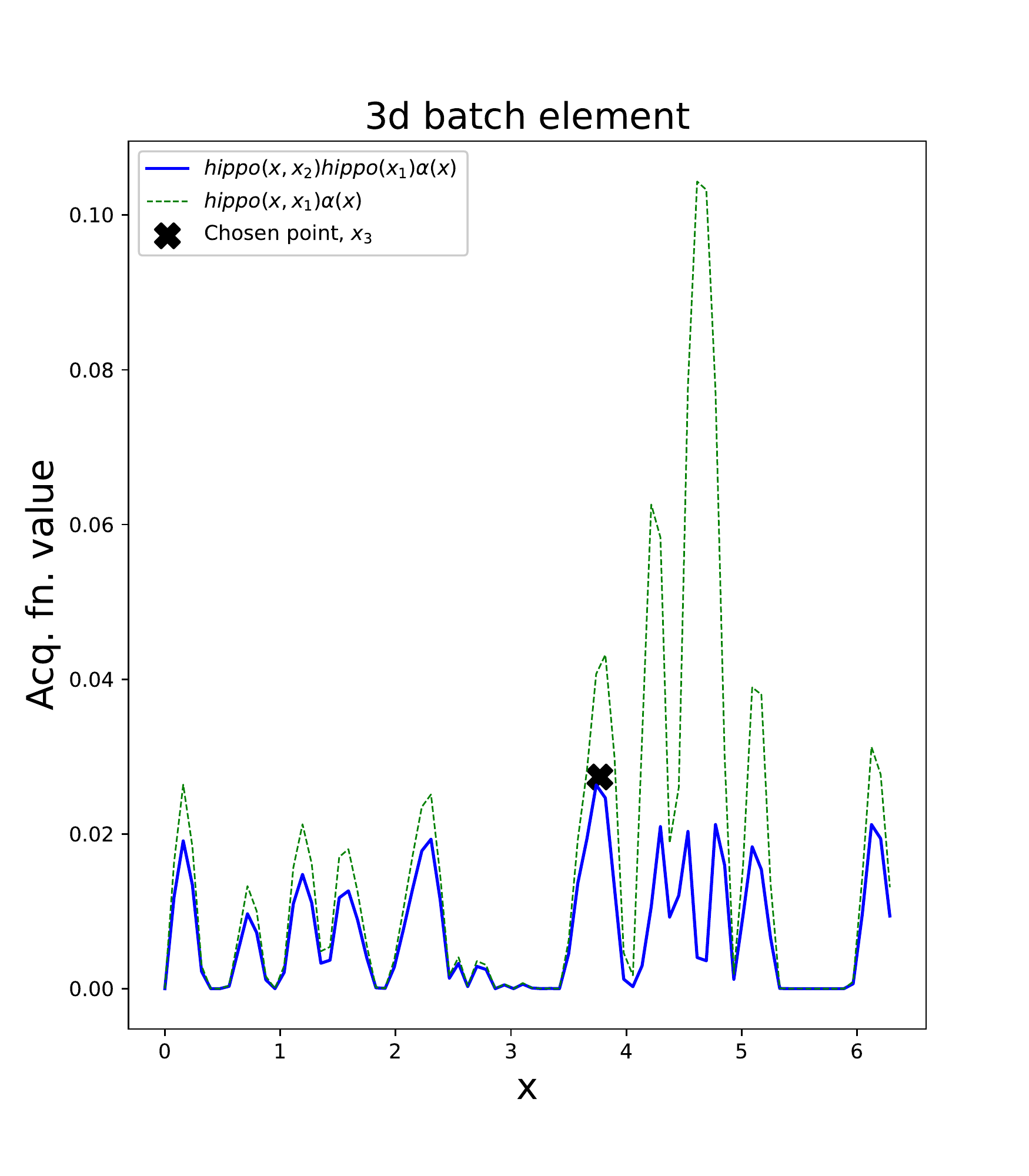}
	\end{subfigure}
	\caption{The process of collecting three consecutive batch points with HIPPO. The first point $x_1$ is collected by maximising the acquisition function $\alpha(x)$ (left). The second point $x_2$ is obtained by multiplying the acquisition function $\alpha(x)$ with the HIPPO penaliser $hippo(x, x_1)$ and maximising the resulting penalised acquisition (center). Third point $x_3$ is obtained by applying HIPPO penalisation to $\alpha(x)$ at both $x_1$ and $x_2$ (right). Dashed green lines on steps 2 and 3 show the acquisition function from the previous step.}
	\vspace{-15pt}
	\label{fig:batch-points}
\end{figure*}

In practice we found HIPPO to be resilient to a particular choice of warping function $w$ as long as it scaled the distance value to $[0, 1)$. For our experiments we used $w(y) = \frac{2}{\pi} \times arctan(y)$. Figure~\ref{fig:batch-points} illustrates the first three steps of the batch collection process with HIPPO for a simple function with 1D input. 




\vspace{-10pt}
\section{Experimental evaluation}
\vspace{-5pt}
\subsection{Benchmark results}
\vspace{-5pt}
This section compares HIPPO with several state-of-the-art methods for multi-objective batch BO on a set of commonly used benchmark functions, considering first modest and then large batch sizes.
As baselines for comparison we use Batch MC, KB and DGEMO. 
To measure the performance of each method we use hypervolume (HV) regret, the difference between the hypervolumes defined by the discovered Pareto fronts and the exact ones. If $HV(\mathcal{P})$ is a function that calculates hypervolume of a given Pareto front $\mathcal{P}$ \citep{wagner2010expected}, then HV regret is understood as $\Delta H = HV(\mathcal{P}_0) - HV(\mathcal{P}_1)$, where $\mathcal{P}_0$ is the exact Pareto front of a given benchmark function (which we calculated with pymoo \citep{pymoo}), and $\mathcal{P}_1$ is the currently discovered Pareto front.

Additionally we measured the time taken for a single optimisation step to show the gain in computational efficiency HIPPO offers compared to other methods. Each optimisation was repeated 10 times.
Batches of size four were used for all experiments, while the number of optimisation steps depended on the problem. We used the \texttt{trieste} library \citep{Berkeley_Trieste_2022} to implement all the experiments\footnote{Experiments code can be found at  \url{https://github.com/apaleyes/hippo}}. We used following benchmark functions: \textit{VLMOP2} \citep{veldhuizen1999multiobjective}, \textit{DTLZ2} \citep{deb2002scalable}, and combination of \textit{Hartmann} and \textit{Ackley} functions \citep{picheny2013benchmark,adorio2005mvf}.

\begin{figure*}
	\newcommand{\scale}{0.32}
	\begin{subfigure}[b]{\scale\textwidth}
		\centering
		\includegraphics[width=\textwidth]{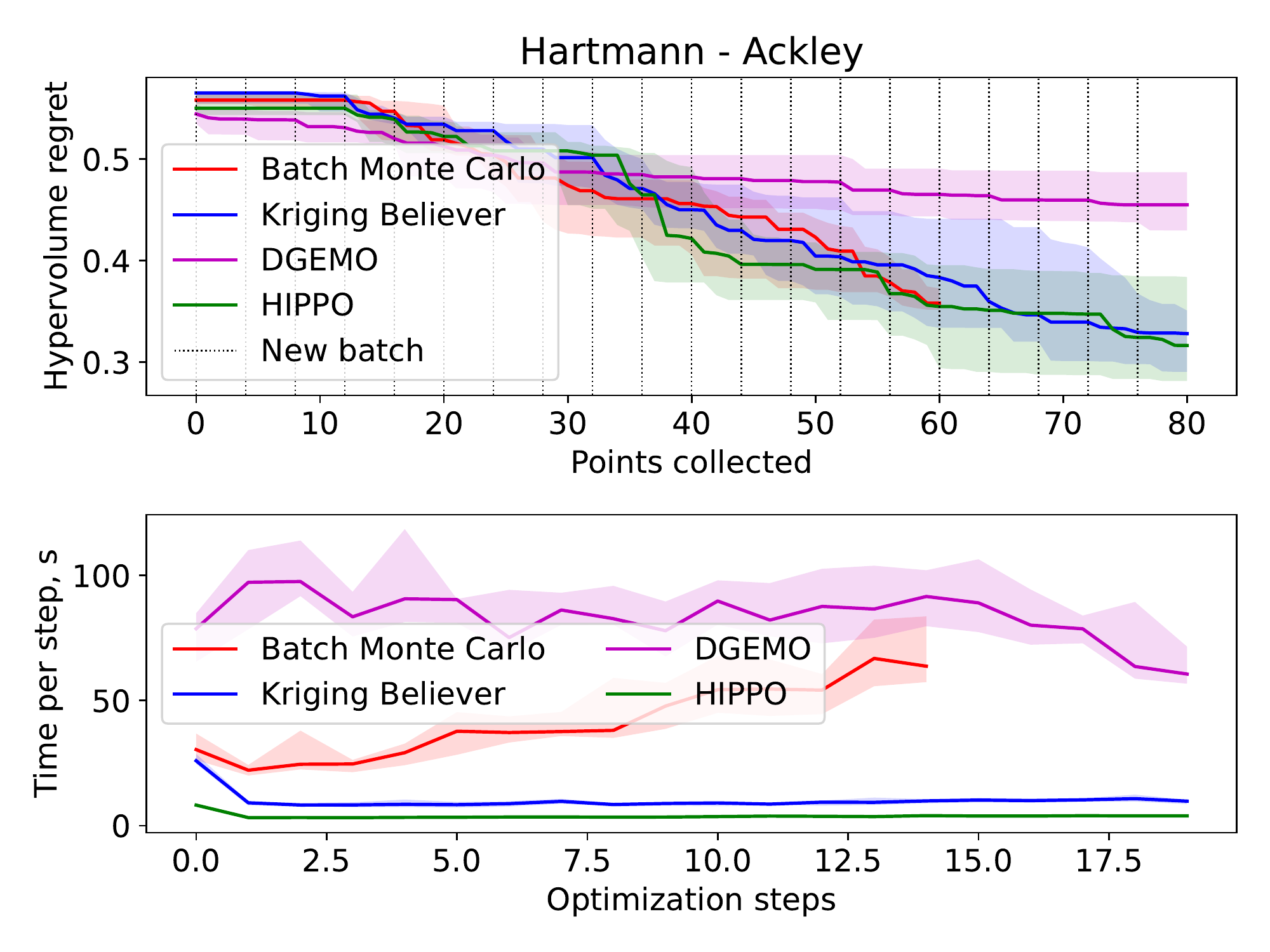}
	\end{subfigure}
	\begin{subfigure}[b]{\scale\textwidth}
	    \centering
		\includegraphics[width=\textwidth]{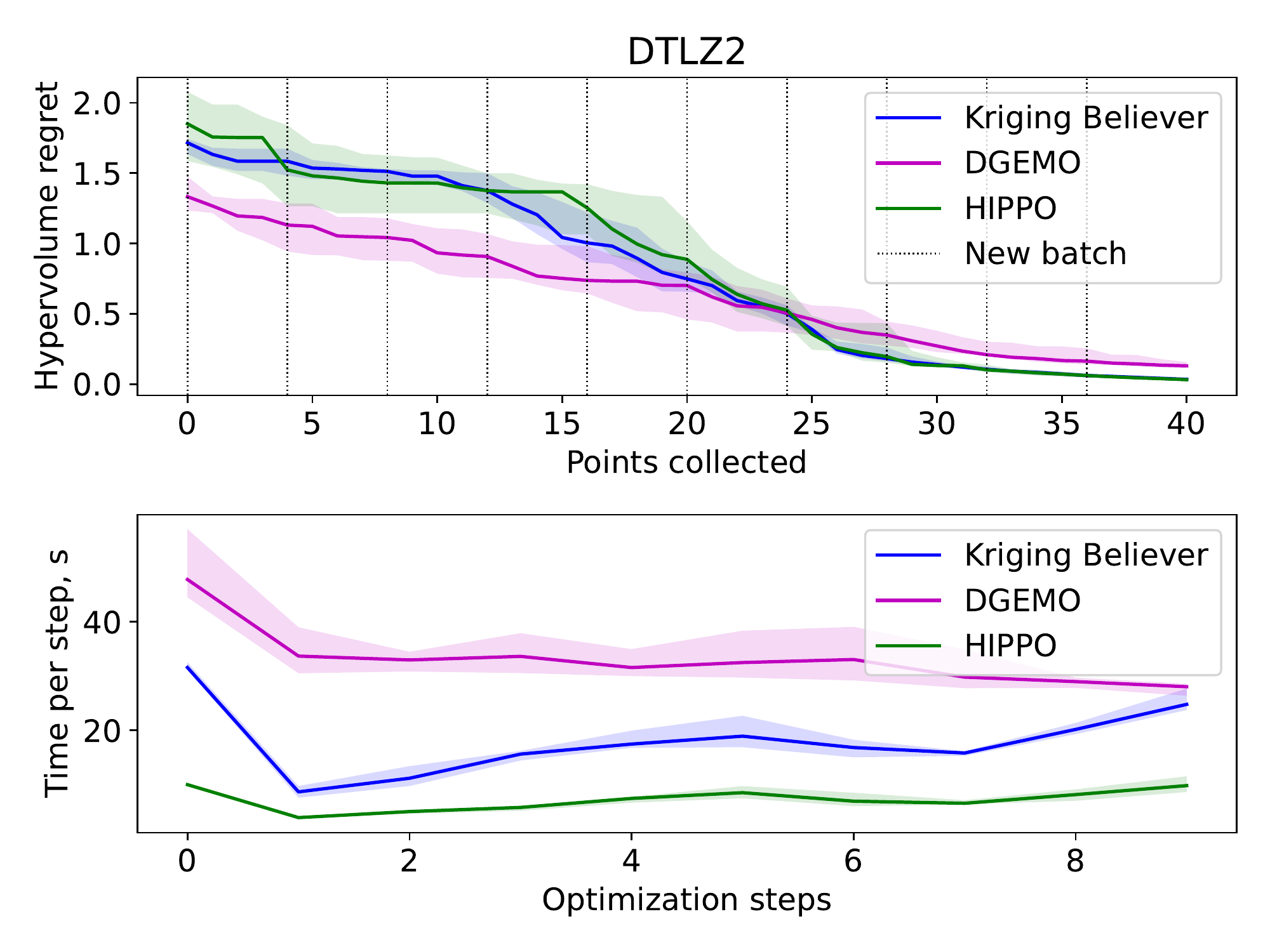}
	\end{subfigure}
	\begin{subfigure}[b]{\scale\textwidth}
	    \centering
		\includegraphics[width=\textwidth]{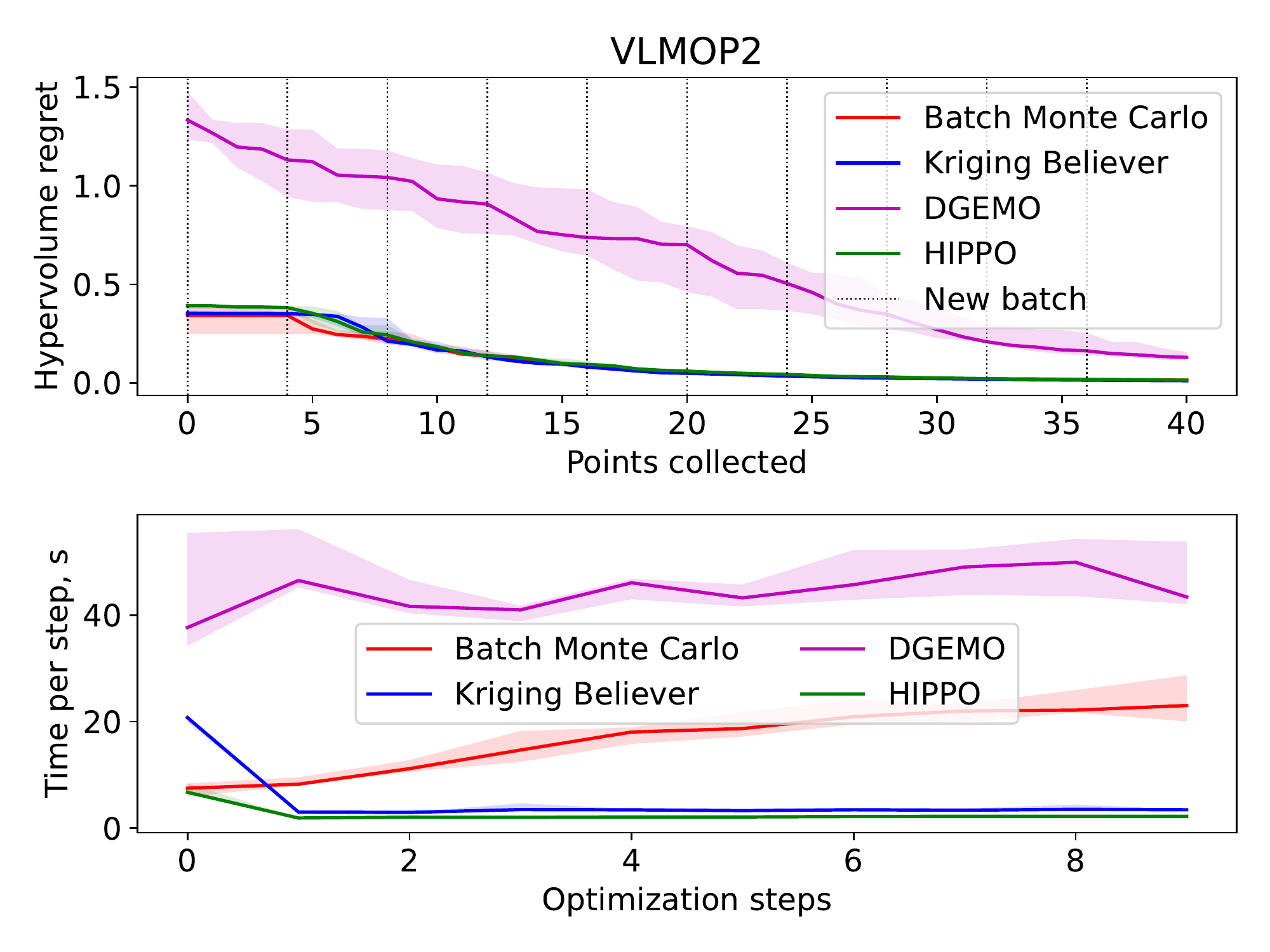}
	\end{subfigure}
	\caption{HV regret (top) and time per optimisation step (bottom) on optimisation benchmark functions. The solid lines show the mean and the shaded areas the $25^{\textrm{th}}$ to $75^{\textrm{th}}$ percentiles envelope. Batch MC performance is not shown for DTLZ2 and last five iterations on Hartmann - Ackley because it was running out of memory in our tests.}
	\vspace{-10pt}
	\label{figure:moo-benchmarks}
\end{figure*}

Results of the experiments are shown on Figure~\ref{figure:moo-benchmarks}. As the HV regret plots show, HIPPO produces Pareto fronts that are on par or better with those produced by other methods. However, as can be seen on the bottom row, even for a modest batch size of 4 HIPPO introduces significantly smaller computational overhead than Batch MC and DGEMO.


\begin{figure*}
	\newcommand{\scale}{0.32}
	\begin{subfigure}[b]{\scale\textwidth}
		\centering
		\includegraphics[width=\textwidth]{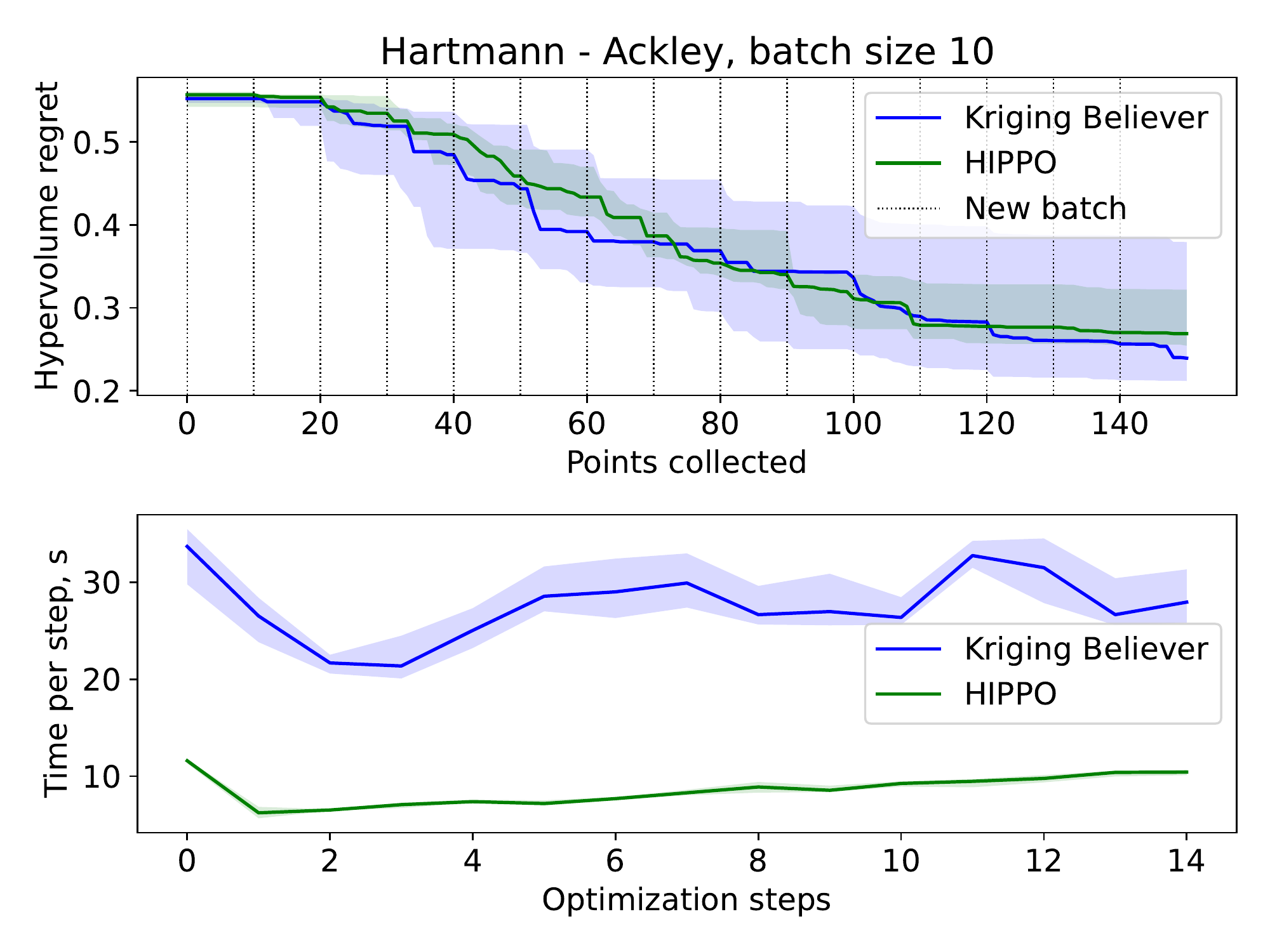}
	\end{subfigure}
	\begin{subfigure}[b]{\scale\textwidth}
	    \centering
		\includegraphics[width=\textwidth]{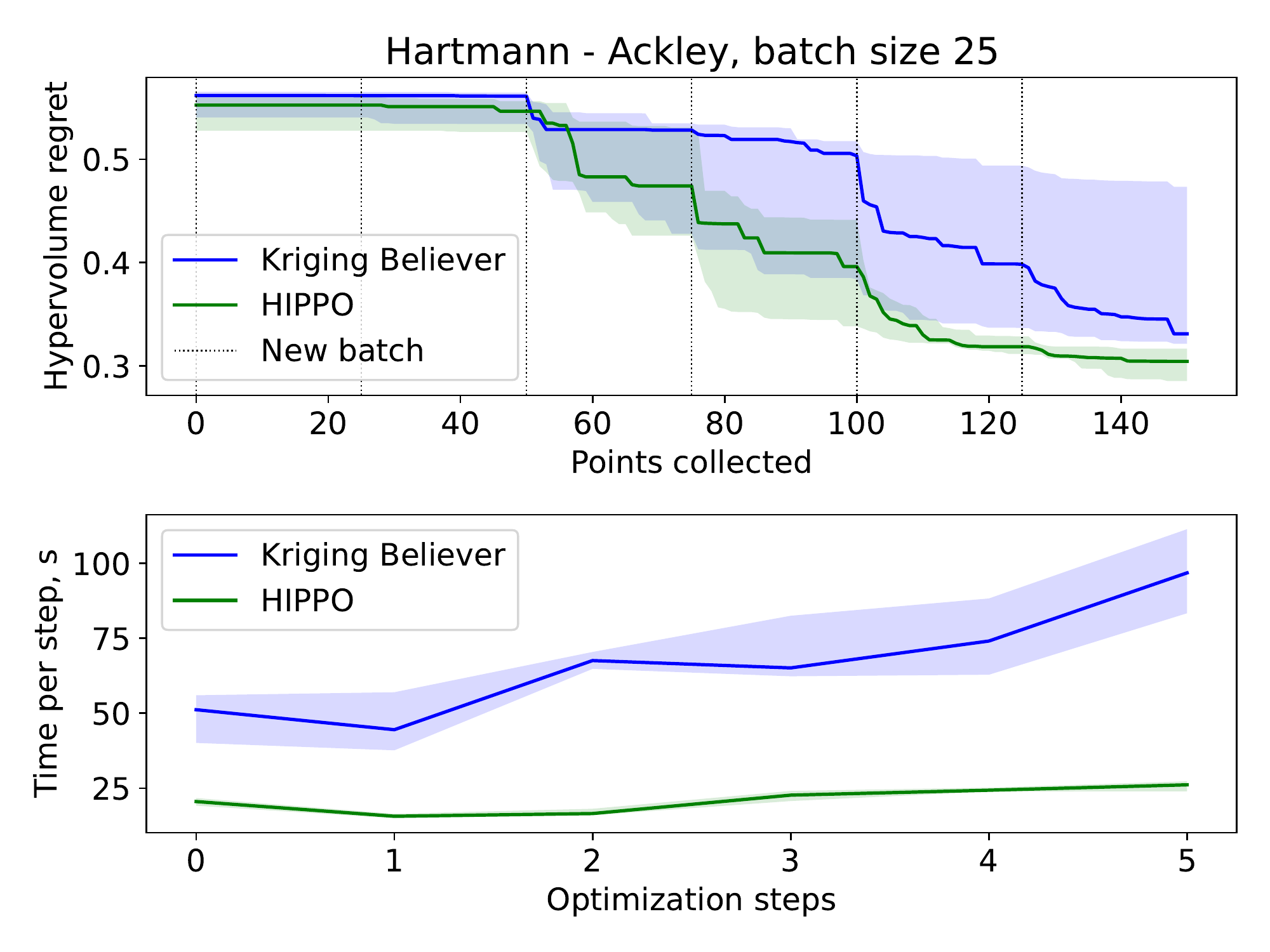}
	\end{subfigure}
	\begin{subfigure}[b]{\scale\textwidth}
		\centering
		\includegraphics[width=\textwidth]{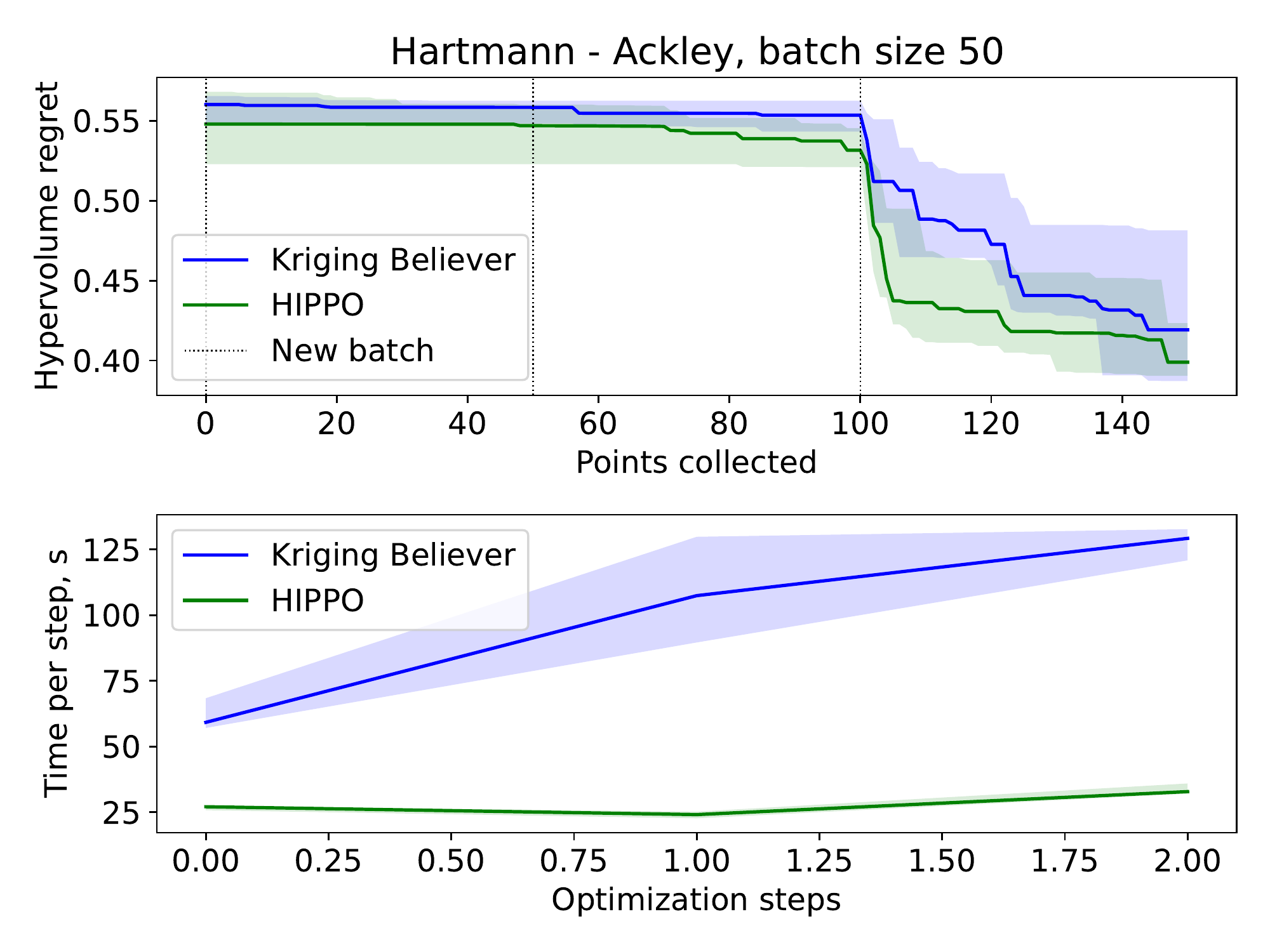}
	\end{subfigure}
	\begin{subfigure}[b]{\scale\textwidth}
		\centering
		\includegraphics[width=\textwidth]{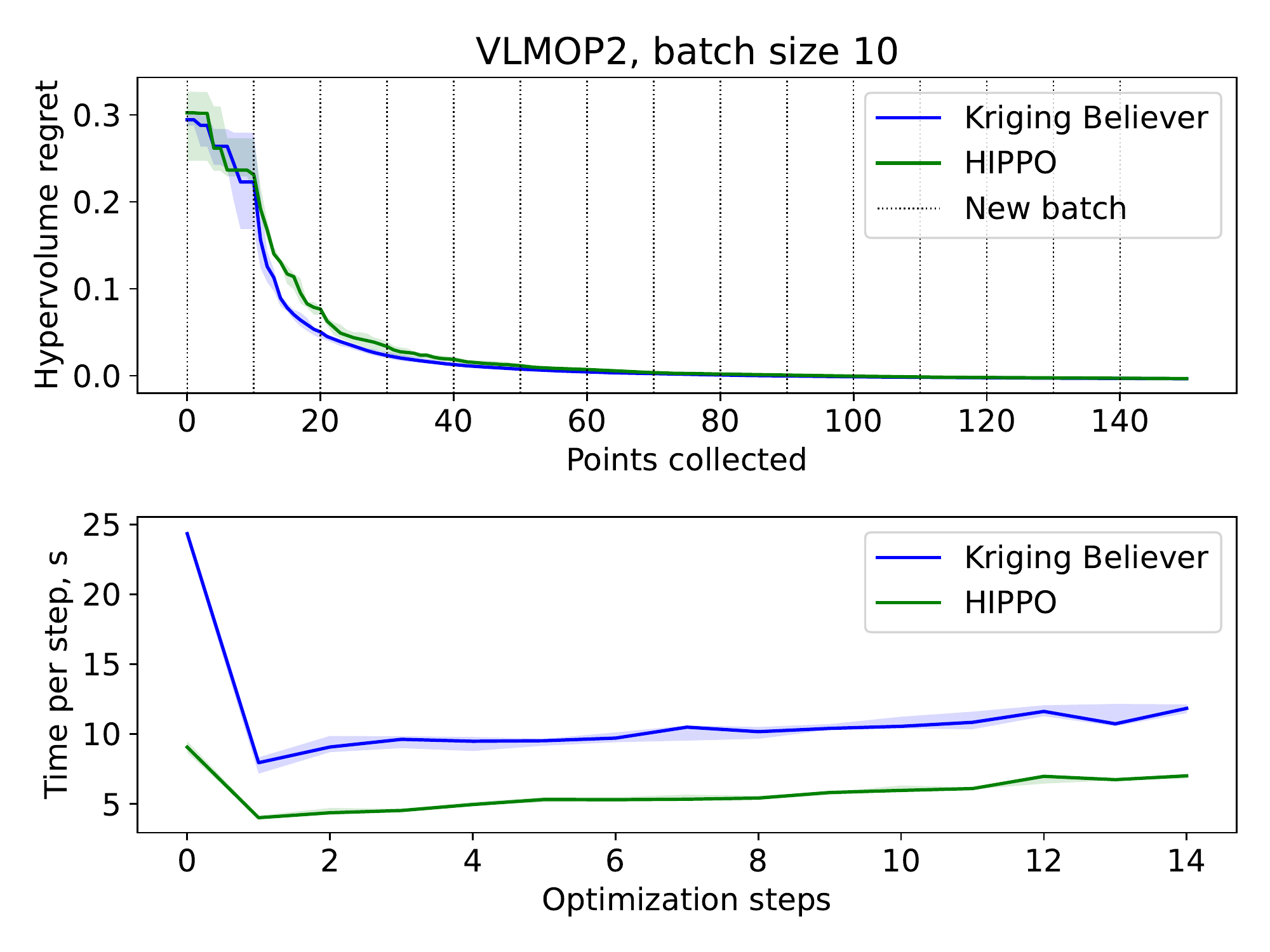}
	\end{subfigure}
	\begin{subfigure}[b]{\scale\textwidth}
	    \centering
		\includegraphics[width=\textwidth]{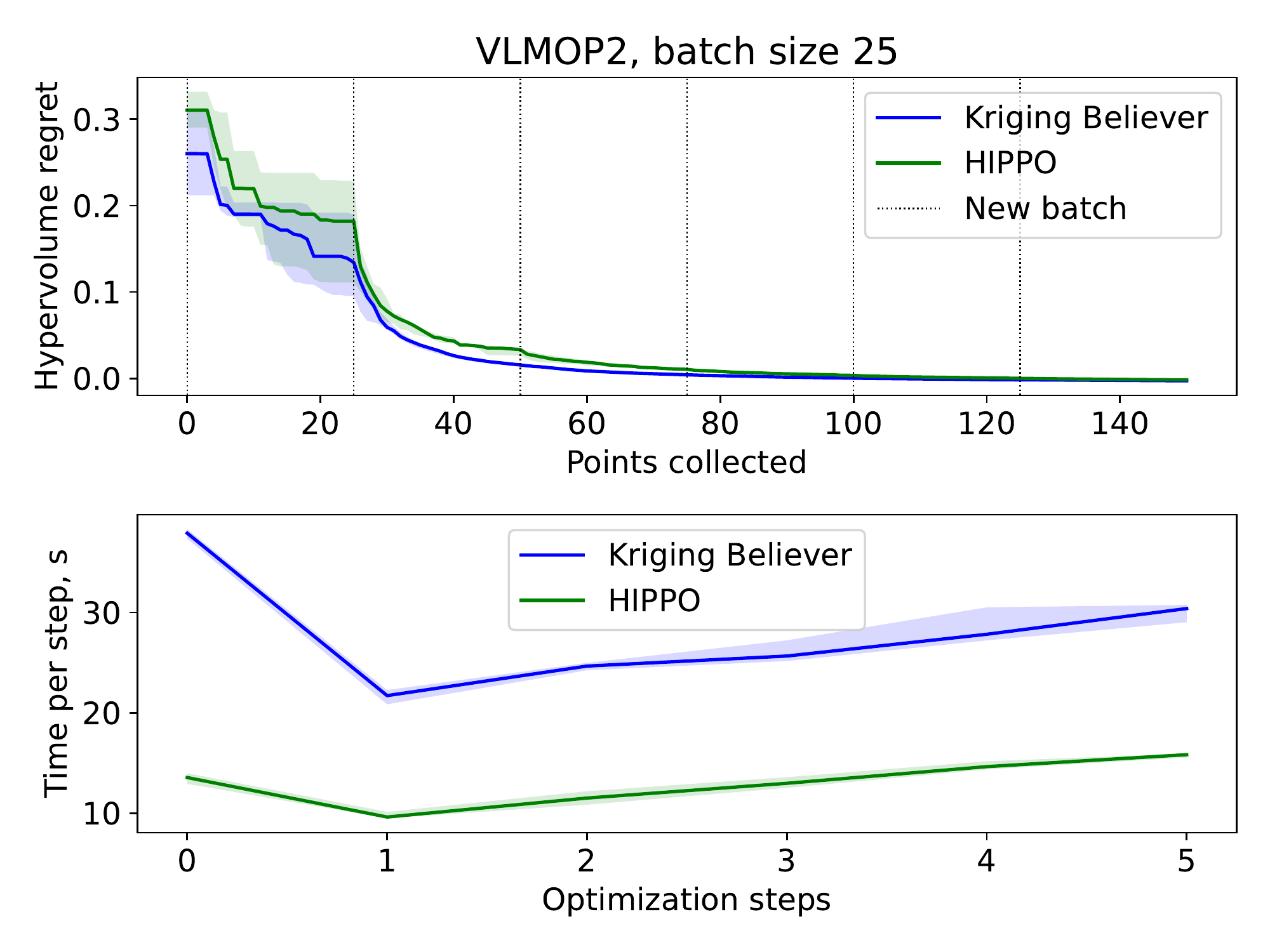}
	\end{subfigure}
	\begin{subfigure}[b]{\scale\textwidth}
		\centering
		\includegraphics[width=\textwidth]{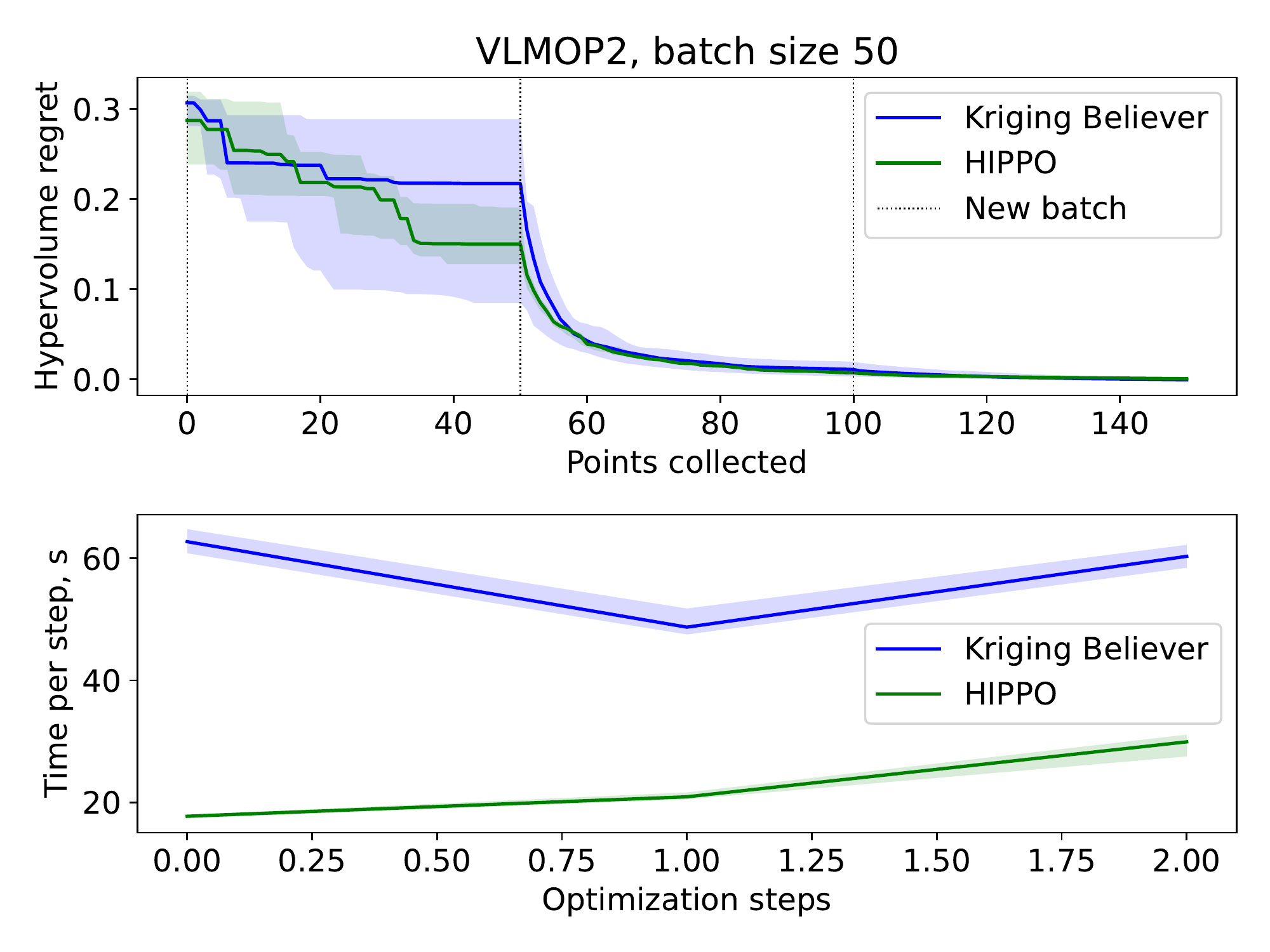}
	\end{subfigure}
	\caption{
	HV regret and time per optimisation step with variable batch sizes. 
	The solid lines show the mean and the shaded areas the $25^{\textrm{th}}$ to $75^{\textrm{th}}$ percentiles envelope. 
	}
	\vspace{-10pt}
	\label{figure:batch-size-experiment}
\end{figure*}

To study scalability of HIPPO, we conducted further experiments with variable batch size. Here we are using \textit{Hartmann - Ackley} and \textit{VLMOP2} as test problems, and \textit{KB} as a baseline, since this method showed the best computational performance of all baselines used in the previous experiment. We conducted the experiment for batch sizes 10, 25 and 50. All experiments were run with the total budget of 150 points, thus yielding 15, 6 and 3 optimisation steps respectively. Experiment results can be found on Figure~\ref{figure:batch-size-experiment}. Again, we can observe very close results in terms of HV regret, however HIPPO has clearly lower computational overhead that grows slowly with the increase in batch size.

\vspace{-10pt}
\subsection{Optimisation of a heat exchanger}
\vspace{-5pt}
We now consider the real-world MOO task of designing an effective yet light-weight automotive heat exchanger (radiator).
The heat exchanger is parameterised with respect to its dimensions and internal geometry. The performance of a particular configuration is given by an expensive-to-evaluate and highly non-linear digital twin based on aero-thermal analysis (a coupling of heat transfer and fluid dynamics).

The input space consists of 9 parameters, with 6 varying continuously between a lower and an upper bound (e.g. radius of component, etc.), as well as 3 taking a finite number of values (e.g. number of modules, etc.).
The two objectives are mass and air-side pressure drop. Additionally, as a minimal performance is requested on two other outputs (heat load and tube-side pressure drop), we also have two inequality constraints. Finally, some configurations are physically impossible or create very unstable conditions which lead to evaluation crashes. We encoded this as an additional binary constraint.

The two objectives and two quantitative constraints are modelled with standard GPR models (with a Matern 5/2 kernel and constant mean function), while the binary constraint is modelled with a variational GP model with Bernoulli likelihood.
As a base acquisition function, we use here EHVI, multiplied by the probability that the three constraints are satisfied \citep[as recommended in ][]{emmerich2006single,shimoyama2013updating}. 

We compare a purely sequential EHVI run with HIPPO using a batch size of 20. Both start with 100 observations (randomly sampled across the input space) and 500 additions. To assess the overall performance of both, a bespoke evolutionary strategy is also ran for a very large number of iterations, which represents the current solutions used in the design of heat exchangers.

\begin{figure*}
\begin{tabular}{ccc}
EHVI & HIPPO &\\
    \includegraphics[trim=5mm 0mm 10mm 10mm, clip, width=.40\textwidth]{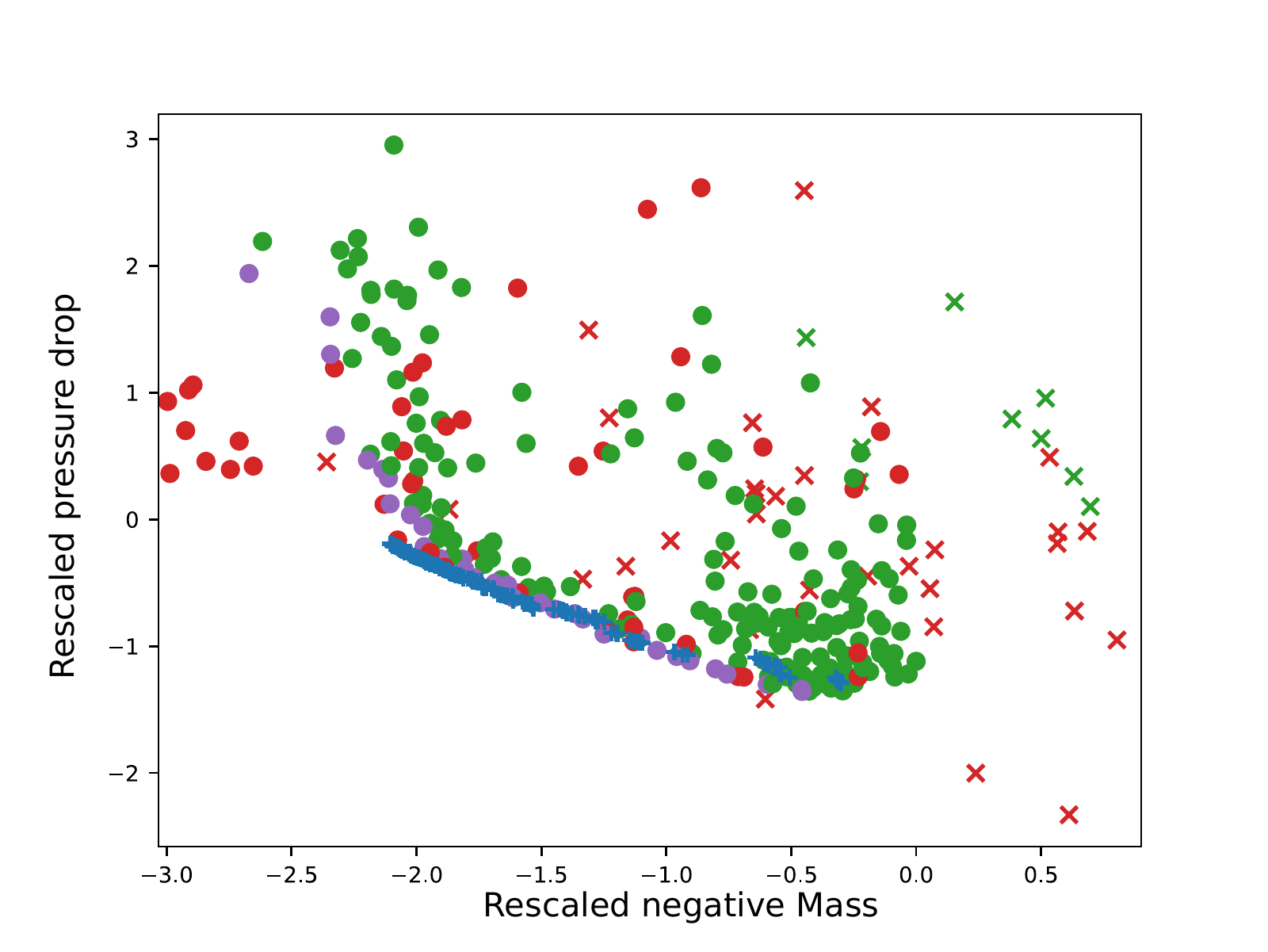} &
    \includegraphics[trim=5mm 0mm 10mm 10mm, clip, width=.40\textwidth]{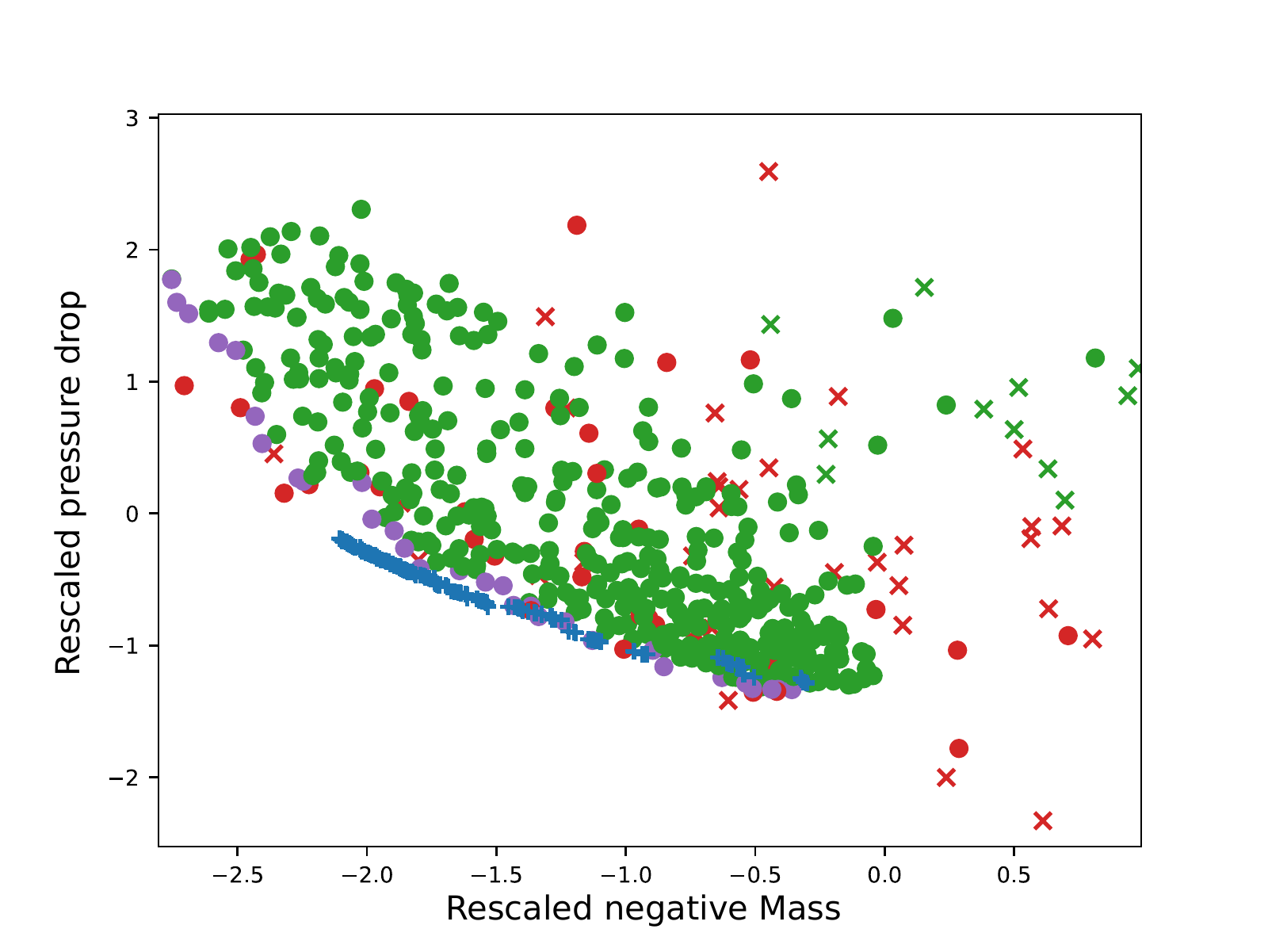} &
    \includegraphics[trim=97mm 40mm 17mm 16mm, clip, width=.2\textwidth]{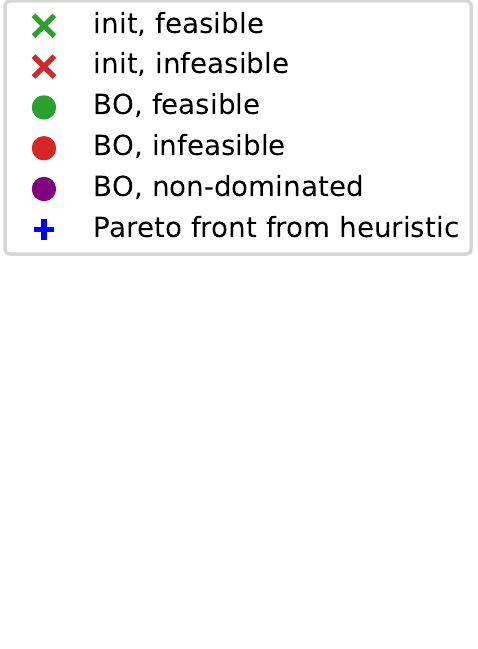}
    \end{tabular}
	\caption{Observation sets in the objective space obtained by EHVI and HIPPO on the heat exchanger problem.
	Both objectives need to be minimised and the utopia point is set as the bottom left corner.}
	\vspace{-10pt}
	\label{figure:reaction-engine}
\end{figure*}

Both Pareto fronts are shown in Figure \ref{figure:reaction-engine}. One can see that the fronts are broadly similar, matching or improving upon the evolutionary baseline on the right of the front and providing a wider range of solutions on the left-most part. Both EHVI and HIPPO returned a balanced set of feasible and non-feasible solutions and a relatively uniform coverage of the Pareto front. However, while EHVI concentrated most of its search on the rightmost part of the front, HIPPO shows a much more balanced search, as expected by its definition. Overall, despite using large batches (20), HIPPO provides comparable performance to EHVI, hence allowing consirable speed-ups of the optimisation pipeline.

\bibliography{references}

\end{document}